\documentclass[11pt,letterpaper]{article}

\usepackage[letterpaper,margin=1in]{geometry}
\usepackage[T1]{fontenc}
\usepackage[utf8]{inputenc}
\usepackage{lmodern}
\usepackage{microtype}
\usepackage{amsmath}
\usepackage{amssymb}
\usepackage{amsthm}
\usepackage{bm}

\usepackage{graphicx}
\graphicspath{{figures/}}
\usepackage{booktabs}
\usepackage{array}
\usepackage{multirow}
\usepackage{tabularx}
\usepackage{caption}
\usepackage{float}

\usepackage{titlesec}
\titleformat{\section}{\normalfont\Large\bfseries}{\thesection}{0.6em}{}
\titleformat{\subsection}{\normalfont\large\bfseries}{\thesubsection}{0.6em}{}
\titleformat{\subsubsection}{\normalfont\normalsize\bfseries}{\thesubsubsection}{0.6em}{}
\titlespacing*{\section}{0pt}{1.6em}{0.8em}
\titlespacing*{\subsection}{0pt}{1.2em}{0.6em}
\titlespacing*{\subsubsection}{0pt}{0.9em}{0.4em}

\usepackage[numbers,sort&compress]{natbib}
\usepackage{xcolor}
\definecolor{linkblue}{RGB}{28,89,158}
\usepackage[colorlinks=true,linkcolor=black,citecolor=linkblue,urlcolor=linkblue]{hyperref}

\newcommand{\kwta}{\operatorname{k\text{-}WTA}}
\newcommand{\softmax}{\operatorname{softmax}}
\newcommand{\onehot}{\operatorname{onehot}}
\newcommand{\normalizeop}{\operatorname{normalize}}
\newcommand{\retrieveop}{\operatorname{retrieve}}
\newcommand{\bigramop}{\operatorname{bigram}}
\newcommand{\errop}{\operatorname{err}}
\newcommand{\logitsop}{\operatorname{logits}}
\newcommand{\retB}{\operatorname{ret}_B}
\newcommand{\retR}{\operatorname{ret}_R}

\title{\vspace{-2em}\textbf{The Art of Not Forgetting
}\vspace{-0.4em}}

\author{
Ashmith Atmuri$^{*}$  \and
Akshay Kumar \and
Yashaswini Rao Bhogarajula \\[0.4em]
\textit{Arkadhi Labs} \\[0.5em]
\small
$^{*}$Corresponding author: \texttt{founder@arkadhi.com} \\[0.3em]
\texttt{ak@arkadhi.com} \quad
\texttt{yashaswini.rao@arkadhi.com}
}

\date{}

\begin{document}
\maketitle

\begin{abstract}
\noindent
We introduce CMP (Cognitive Memory Primitive), an architecture that represents inputs as sparse relational codes, stores them in a two-tier competitive memory, and learns entirely through local, gradient-free updates, with no backpropagation anywhere in the network. We use this architecture to test a specific hypothesis: that catastrophic forgetting, usually treated as a training-time defect to be patched with replay or regularization, is instead a structural consequence of how backpropagation assigns credit, and that a learning rule which is local and sparse by construction should resist it without a patch. On a controlled domain-incremental protocol across 15 text domains, three-seed replicated, CMP's backward transfer is 15--19$\times$ better than a matched-size Transformer trained with online EWC, and the result survives a domain-order control (reported as a range, $+0.24$ to $+0.44$, rather than a single figure). We report this alongside a real, substantial accuracy gap versus the Transformer baseline, a null result on a recognized vision benchmark, and a diagnosed, unresolved failure attempting to combine this architecture with a separate mechanism that improves raw accuracy, disclosed because an honest negative result is more useful than an omitted one. The central claim is narrow and falsifiable: local, sparse, non-backpropagation learning measurably resists catastrophic forgetting better than backpropagation with its standard fix, under conditions we state precisely.
\end{abstract}

\vspace{0.3em}

\section{Introduction}
\label{sec:introduction}

Catastrophic forgetting, the tendency of a network trained sequentially on new data to overwrite what it learned on old data, is typically treated as a training-time problem to be patched: replay buffers, regularization terms like EWC~\citep{kirkpatrick2017overcoming}, or architectural additions like Progressive Networks~\citep{rusu2016progressive}. Each of these treats forgetting as incidental to backpropagation, something to be corrected after the fact. We investigate a different premise: that forgetting may instead be a structural consequence of how backpropagation itself assigns credit, globally, densely, touching every weight in proportion to a single scalar loss, and that a learning rule which is local, sparse, and gradient-free by construction may resist it for architectural reasons, not because of an added correction term.

We test this premise directly. CMP (Cognitive Memory Primitive) combines an established binding mechanism~\citep{plate1995holographic,pollack1990recursive} with a sparse, competitive, content-addressed memory in the tradition of Kanerva~\citep{kanerva1988sparse} and modern Hopfield networks~\citep{ramsauer2021hopfield}, a predictive-coding cascade~\citep{rao1999predictive,whittington2017approximation}, and a local delta-rule readout; no component of which is individually novel (Section 3). What we add is a specific combination, including a gradient-free, movement-based plasticity throttle (Section 3.7) in place of Fisher-information-based importance, and an empirical test of whether that combination measurably resists forgetting relative to backpropagation with its standard fix.

\subsubsection*{We state plainly, here, what this paper does not claim.}
It does not claim CMP is more accurate than a Transformer; it is not, by a substantial margin (Section 6). It does not claim general superiority over attention-based architectures. It claims a single, narrower, falsifiable result: on a controlled domain-incremental protocol, with matched architecture size and matched data budget, CMP's backward transfer is substantially better than a Transformer trained with online EWC, and the effect survives a domain-order control and three-seed replication.

\subsection{Contributions}
\label{subsec:contributions}
\begin{itemize}
  \item A domain-incremental continual-learning result across 15 real-world text domains, three-seed replicated, showing CMP forgets 15--19$\times$ less than a matched Transformer under naive fine-tuning and online EWC (Section 4).
  \item A gradient-free importance signal for plasticity throttling, substituting raw weight-movement magnitude for EWC's Fisher information, motivated by the absence of a backpropagated gradient to square in a local-learning system.
  \item A domain-order control showing the effect is real but sequence-dependent, reported as a range rather than a single figure.
  \item An honest accounting of what does not yet work: a real accuracy gap, a null result on Split-MNIST, and a diagnosed, unresolved failure attempting to combine this architecture with a separate, higher-accuracy local-learning mechanism (Appendix D).
\end{itemize}

\section{Related Work}
\label{sec:related}

\paragraph{Sparse coding.}
Olshausen and Field~\citep{olshausen1996emergence} showed that a sparse coding objective applied to natural images recovers receptive fields resembling primary visual cortex, establishing sparsity as a biologically-grounded representational choice. CMP's k-winners-take-all sparsification throughout (Section 3.1) inherits this motivation directly.

\paragraph{Binding and compositional representation.}
The multiplicative binding operation (Section 3.1) belongs to the Vector Symbolic Architecture family: Pollack's Recursive Distributed Representations~\citep{pollack1990recursive} and Plate's Holographic Reduced Representations~\citep{plate1995holographic} both address representing compositional structure in fixed-width distributed vectors via binding operations. Neither the binding operation nor its use of superposition is new; what is combined with it (Section 3.2--3.7) is specific to this work.

\paragraph{Associative memory.}
Kanerva's Sparse Distributed Memory~\citep{kanerva1988sparse} is the direct ancestor of CMP's competitive, content-addressed memory (Section 3.2). Ramsauer et al.~\citep{ramsauer2021hopfield} establish that modern Hopfield retrieval is mathematically continuous with self-attention; we position CMP's memory as a sibling within that same associative-memory family, distinguished by sparsity and recurrent, one-cue-at-a-time retrieval rather than dense, simultaneous all-pairs lookup, not as a departure from it.

\paragraph{Predictive coding.}
Rao and Ballard~\citep{rao1999predictive} propose predictive coding as a functional account of cortical feedback; Whittington and Bogacz~\citep{whittington2017approximation} show a predictive-coding network can approximate error backpropagation using only local Hebbian updates. CMP's predictive-coding cascade (Section 3.4) uses the mechanism for its stated purpose, local, self-supervised top-down prediction, rather than as an approximation to backpropagation.

\paragraph{Complementary learning systems.}
McClelland, McNaughton, and O'Reilly~\citep{mcclelland1995why} argue the hippocampus and neocortex form complementary systems precisely to avoid the interference that a single fast-learning system would suffer: the direct theoretical motivation for pairing a sparse, fast-writing memory with a slower, distributed readout in CMP, and for expecting that pairing to resist forgetting for structural reasons.

\paragraph{Continual learning.}
EWC~\citep{kirkpatrick2017overcoming} regularises parameter movement by Fisher-information-weighted importance; Synaptic Intelligence~\citep{zenke2017continual} computes importance online from the training trajectory; Progress \& Compress~\citep{schwarz2018progress} introduces the online-EWC variant used as our baseline (Section 4.3). Progressive Networks~\citep{rusu2016progressive}, instead, avoid architectural forgetting by freezing old columns and adding new ones per task. iCaRL~\citep{rebuffi2017icarl} represents the replay-based family, noted but not compared directly. Lopez-Paz and Ranzato~\citep{lopezpaz2017gradient} define the BWT/FWT metrics used throughout Section 4, adapted here to a bits-per-byte formulation with an explicitly stated sign convention.

\paragraph{Non-attention sequence architectures.}
Mamba~\citep{gu2023mamba} and related state-space models achieve strong long-context efficiency without attention, but, to our knowledge, and consistent with our own accuracy results (Section 6), no non-attention architecture has cleanly matched Transformer accuracy at matched parameters and compute on short-context tasks. We note this as precedent for reporting a real accuracy gap alongside a different, real contribution.

\paragraph{Byte-level language modelling.}
The Byte Latent Transformer~\citep{pagnoni2025byte} allocates compute dynamically based on next-byte entropy, operating, like CMP, directly on bytes rather than a learned subword vocabulary. We note the shared byte-level premise as context for Section 4's bits-per-byte metric.

\section{Architecture}
\label{sec:architecture}

Before the equations, the shape of the whole thing: a byte pair is bound into one sparse code (3.1); that code is written into a two-tier memory it can later be retrieved from (3.2); a second, coarser code is formed above it and used to predict the fine code, with the prediction error kept as a signal (3.3--3.4); every term produced so far, the code itself, the memory retrieval, the coarse code, and the error, is summed at a single readout layer to produce a next-byte prediction (3.5); that readout is updated with a local rule that needs no backpropagation through any of the preceding steps (3.6); and a separate, gradient-free importance signal throttles how fast each weight is allowed to change, which is the one piece of this pipeline that is not an established technique on its own (3.7). Section 3.8 then compares this whole design against self-attention directly, on the specific axes that motivate building it this way instead.

Figure 1 provides a high-level view of the complete CMP pipeline. The following subsections explain each component in order.

\begin{figure}[H]
  \centering
  \includegraphics[width=0.55\linewidth]{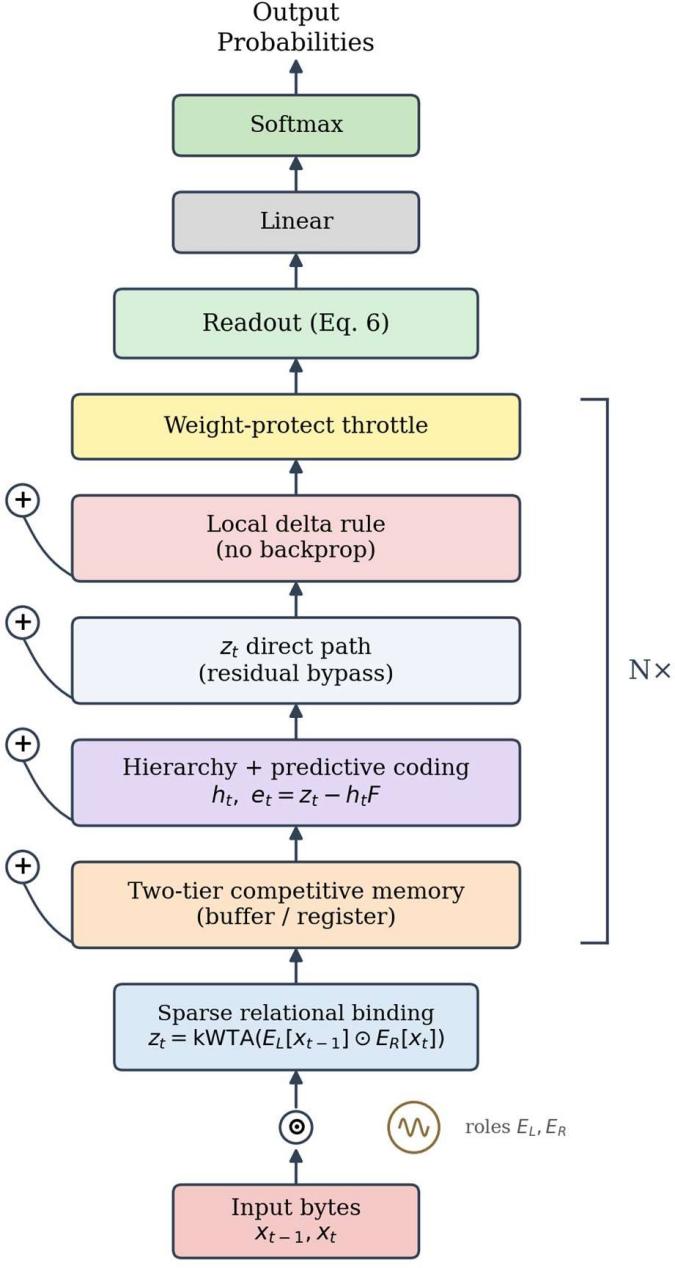}
  \caption{CMP architecture: binding $\rightarrow$ memory $\rightarrow$ hierarchy/predictive coding $\rightarrow$ readout $\rightarrow$ local learning $\rightarrow$ weight-protect.}
  \label{fig:cmp-pipeline}
\end{figure}

We describe the architecture in the same order Vaswani et al.~\citep{vaswani2017attention} describe self-attention: the core binding operation first, then how it composes into memory, hierarchy, and a learning rule. A small version of this figure, with the relevant piece highlighted, appears at the start of each subsection below so it is always clear which part of the pipeline is being discussed.

\subsection{Sparse relational binding}
\label{subsec:binding}

\begin{center}
  \includegraphics[width=0.42\linewidth]{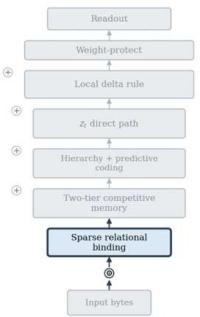}
\end{center}

The problem this step solves: turn two consecutive raw bytes into a single fixed-size object that later steps can store, compare, and combine, without a per-pair lookup table that would grow with vocabulary size, and without keeping the raw sequence around the way self-attention does. Input: the previous byte $x_{t-1}$ and the current byte $x_t$. Output: a single sparse vector $z_t$. Two frozen random embedding matrices $E_L$, $E_R$ map each byte to a left- and right-role vector; the core operation binds them via elementwise product, followed by k-winners-take-all sparsification and renormalization:

\begin{equation}
\label{eq:binding}
z_t \;=\; \normalizeop\!\bigl(\kwta_k\!\bigl(E_L[x_{t-1}] \;\odot\; E_R[x_t]\bigr)\bigr)
\end{equation}

$z_t$ is an $r$-dimensional vector, the relational code at position $t$: CMP's basic unit of representation, analogous to the query/key/value vectors in self-attention but produced by a single multiplicative binding rather than a learned linear projection followed by a dot-product similarity search over the full sequence. $E_L$ and $E_R$ are fixed at initialization, not trained, a deliberate design choice revisited as a limitation in Section 6. Everything from here on operates on $z_t$: the next two subsections are two different things done with it, storage and coarsening, not two alternatives to it.

\subsection{Two-tier competitive memory showing write decision, promotion, and retrieval.}
\label{subsec:memory}

\begin{center}
  \includegraphics[width=0.42\linewidth]{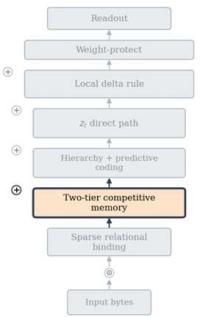}
\end{center}

The problem this step solves: binding alone only encodes local, byte-pair context, and a system with no persistent store has no way to reference something it saw many steps ago without a dense weight matrix that global backpropagation can overwrite wholesale. Input: the current relational code $z_t$, used as a query. Output: a retrieved memory vector, a weighted combination of whatever stored codes resemble $z_t$. A buffer $B$ (fast, short-term) and a register $R$ (slow, long-term) store relational codes via competitive, content-addressed writes. Retrieval for a query $z_t$ is a similarity-weighted read over slots:

\begin{equation}
\label{eq:retrieval}
\retrieveop(M, z_t) \;=\; \sum_{i=1}^{|M|} \softmax_i\!\bigl(\tau\,\hat{z}_t \cdot \hat{m}_i\bigr)\, m_i
\end{equation}

Writes are competitive: a slot updates only if its similarity to $z_t$ exceeds a calibrated match threshold (Eq. 3), so a slot only fires on a genuine match, not baseline overlap. $n_{\text{buffer}}$, $n_{\text{register}}$, $k$ are all derived from $r$ by a fixed ratio (Appendix A), not set independently, a limitation discussed in Section 6.

\begin{equation}
\label{eq:write}
\theta_{\text{match}}(r, k, s) \;=\; 0.86 \cdot \mathbb{E}\!\left[\max_{i \neq j}\, \hat{v}_i \cdot \hat{v}_j\right]
\end{equation}

\begin{figure}[H]
  \centering
  \includegraphics[width=0.7\linewidth]{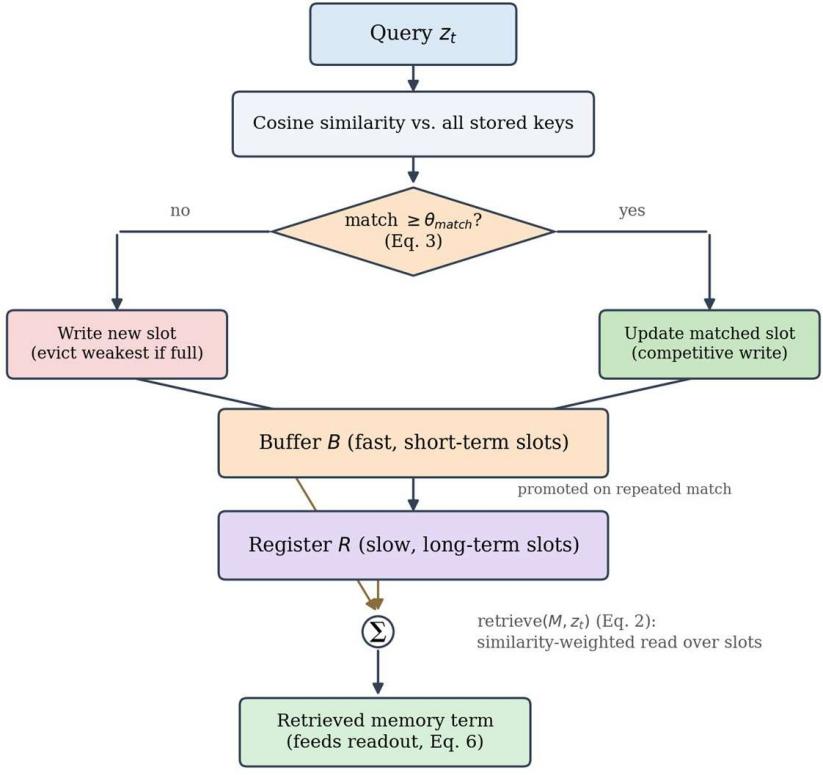}
  \caption{Two-tier competitive memory: the write-side match decision, buffer-to-register promotion, and the retrieval-side merge.}
  \label{fig:memory}
\end{figure}

The takeaway: writes are gated by the same match threshold that governs retrieval, so nothing enters memory on a weak or coincidental overlap, and only codes that survive repeated matching earn promotion from the fast buffer to the slow, long-term register.

Binding and memory both operate at the single-byte-pair scale; the next two subsections ask whether coarser structure, something closer to words than bytes, adds a useful signal on top.

\subsection{Hierarchy}
\label{subsec:hierarchy}

\begin{center}
  \includegraphics[width=0.42\linewidth]{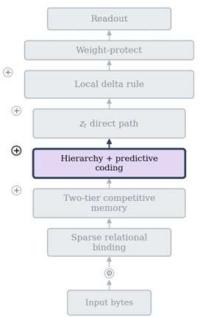}
\end{center}

The problem this step solves: byte-pair binding is too fine-grained to represent word- or phrase-level structure on its own, and adding that would normally mean a second, separately-trained encoder. Input: two consecutive relational codes, $z_{t-1}$ and $z_t$. Output: $h_t$, a second, coarser code, formed by binding consecutive relational codes through their own (frozen) mixing matrices $W_{L2}$, $W_{R2}$:

\begin{equation}
\label{eq:hierarchy}
h_t \;=\; \normalizeop\!\bigl(\kwta_k\!\bigl((z_{t-1}\, W_{L2}) \;\odot\; (z_t\, W_{R2})\bigr)\bigr)
\end{equation}

$h_t$ operates at a coarser temporal granularity than $z_t$ (empirically, word-scale rather than byte-scale): a second, independent zoom level, the CMP analogue of running multiple attention heads at different receptive-field scales, not more of the same binding. Once this coarser code exists, it can be used for more than storage, it can try to predict the fine code, which is exactly what the next step does.

\subsection{Predictive coding}
\label{subsec:predictive}

\begin{center}
  \includegraphics[width=0.42\linewidth]{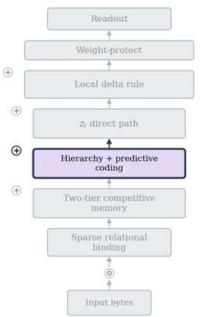}
\end{center}

The problem this step solves: not every byte is equally predictable from its coarse context, and how surprising a byte is, given what the coarser code expected, is itself a useful signal, one that would otherwise require a separate, trained error head. Input: the coarse code $h_t$ and the fine code $z_t$. Output: $e_t$, a sparse residual term. The coarse code $h_t$ predicts the fine code $z_t$ through a locally-trained matrix $F$, and the residual becomes an additional sparse context term:

\begin{equation}
\label{eq:predictive}
\hat{z}_t \;=\; h_t\, F, \qquad e_t \;=\; \normalizeop\!\bigl(\kwta_k(z_t - \hat{z}_t)\bigr)
\end{equation}

$F$ is trained by its own local reconstruction rule and never receives gradient signal from the downstream classification error: the top-down prediction is self-supervised, structurally isolated from the task loss. At this point four terms exist, $z_t$, the memory retrieval, $h_t$, and $e_t$; readout is where all four get combined into an actual prediction.

\subsection{Readout}
\label{subsec:readout}

\begin{center}
  \includegraphics[width=0.42\linewidth]{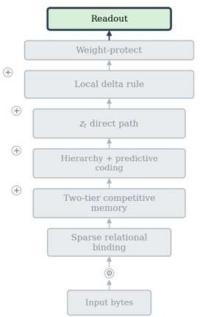}
\end{center}

The problem this step solves: every term produced above needs to become one thing, next-byte logits, without a nonlinear MLP standing between them and the loss, since a local learning rule cannot easily credit-assign through one. Input: $z_t$, the retrieved memory term, $h_t$, $e_t$, and a set of leaky-integrator context terms (Appendix A). Output: logits over 256 byte values. All terms above are linearly combined and added to a bigram table $W_h$:

\begin{equation}
\label{eq:readout}
\logitsop_t \;=\; W_h^{\top}\,[x_t] \;+\; b \;+\; z_t\, W_0^{\top} \;+\; \retB\, W_B^{\top} \;+\; \retR\, W_R^{\top} \;+\; h_t\, W_{h_2}^{\top} \;+\; e_t\, W_{pc}^{\top} \;+\; \sum_{\alpha} c_t^{(\alpha)}\, W_{s_\alpha}^{\top}
\end{equation}

The readout is nothing more than a set of weights, $W_h$, $W_0$, $W_B$, $W_R$, $W_{h_2}$, $W_{pc}$, $W_{s_\alpha}$, summed. The next question is how those weights get updated without backpropagation.

\subsection{Local learning rule}
\label{subsec:local-rule}

\begin{center}
  \includegraphics[width=0.42\linewidth]{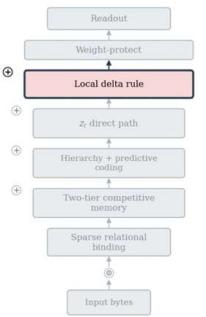}
\end{center}

The problem this step solves: with no backpropagation, there is no chain rule connecting the readout error back through binding, memory, and hierarchy, so every readout weight must be updated using only information available locally, at the readout, on this step. Input: the softmax error at the readout. Output: an updated readout weight, one hop away from the error, no further. Every readout weight is updated by a one-step delta rule using the softmax error, with no backpropagation through Eq. 6's upstream terms:

\begin{align}
\label{eq:delta}
\errop_t &\;=\; \onehot(y_t) \;-\; \softmax(\logitsop_t) \\[0.2em]
\label{eq:update}
\Delta W_{\bullet} &\;=\; \frac{\eta}{N}\sum_{t} \errop_t^{\top}\, \phi_{\bullet}(t)
\end{align}

A plain delta rule, on its own, updates every weight equally hard regardless of which domain is currently training, which is itself a route back to forgetting: weight-protect, next, is the fix for that specific failure mode.

\subsection{Weight-protect: the paper's one concrete contribution}
\label{subsec:weight-protect}

\begin{center}
  \includegraphics[width=0.42\linewidth]{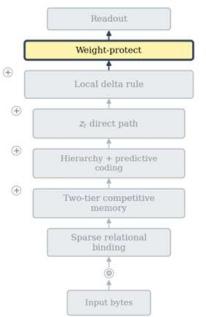}
\end{center}

The problem this step solves: every mechanism described so far (3.1--3.6) treats every domain identically, so a plain delta rule will happily overwrite weights that were important for a domain trained many steps ago. Input: the readout weights' movement, tracked across training. Output: a per-tensor importance score that throttles future updates. After training on domain $i$, the mean absolute movement of each readout tensor is accumulated as an importance score, and future updates are scaled inversely to it:

\begin{align}
\label{eq:importance}
I_{\bullet} &\;\leftarrow\; I_{\bullet} \;+\; \mathbb{E}\!\bigl[\,\bigl|\,W_{\bullet}^{(i)} \;-\; W_{\bullet}^{(i-1)}\,\bigr|\,\bigr] \\[0.2em]
\label{eq:throttle}
\Delta W_{\bullet} &\;\leftarrow\; \frac{1}{1 + \lambda\, I_{\bullet}}\; \Delta W_{\bullet}
\end{align}

\subsubsection*{This is the mechanism the paper's central claim rests on. It is deliberately not Fisher-information-based:}

EWC's importance signal is the expected squared gradient of the log-likelihood, which presumes a backpropagated gradient to square. CMP has no such gradient by construction, so Eq. 9 substitutes raw weight-movement magnitude, a cheaper, gradient-free importance heuristic, and, to our knowledge, not previously used as the basis for an EWC-style throttling rule. We state this as the paper's one concrete, own mechanism-level formalization, not a claim of new mathematics: the binding operation, memory, and delta rule are each established techniques (Section 2); Eq. 9--10 is the specific combination novel to this work.

\subsection{Why sparse relational binding, not self-attention}
\label{subsec:vs-attention}

With all seven pieces now on the table (Figure 1), one question remains before turning to results: why build the core operation this way instead of with self-attention.

\begin{table}[H]
  \centering
  \caption{Complexity and locality comparison, in the spirit of Vaswani et al.~\citep{vaswani2017attention}, Table 1.}
  \label{tab:vs-attention}
  \small
  \begin{tabular}{@{}lll@{}}
    \toprule
    \textbf{Axis} & \textbf{Self-attention} & \textbf{Sparse relational binding} \\
    \midrule
    Compute / step   & $O(n^2)$, all-pairs           & $O(1)$, fixed-size memory \\
    Learning signal  & Backprop, global              & Local delta rule, one-hop \\
    Memory           & None, recomputed each pass    & Explicit, persistent, sparse \\
    \bottomrule
  \end{tabular}
\end{table}

\subsubsection*{The takeaway: sparse relational binding trades self-attention's global, quadratic lookup for a fixed-size, local operation that a gradient-free rule can actually credit-assign through.}

Ramsauer et al.~\citep{ramsauer2021hopfield} show modern Hopfield retrieval is mathematically continuous with self-attention; CMP's memory is a sibling in that same associative-memory family, not a departure from it. What distinguishes it structurally is sparsity and one-cue-at-a-time recurrent retrieval rather than dense, simultaneous all-pairs lookup: the property this paper's central experiment, described next, tests for a downstream consequence.

\section{Continual Learning Protocol and Results}
\label{sec:experiments}

The experiments below build on each other in a fixed order, each one closing a specific hole the previous one leaves open: first confirm the model is learning anything at all (4.2), since a forgetting metric on a model with nothing to forget is meaningless; then confirm the central mechanism's effect replicates rather than being a single lucky run (4.3); then confirm it beats the standard fix at matched capacity and budget, not just a weak baseline (4.4); then confirm the effect is not an artifact of one convenient domain ordering (4.5); then check whether any of it survives outside text at all (4.6).

\subsection{Protocol}
\label{subsec:protocol}

We adopt a domain-incremental continual-learning protocol: a single model is trained sequentially on a fixed ordering of $T$ domains, with no replay of earlier domains' data and no task identifier available at any point. For each domain $i$, the model trains for a fixed number of steps, is evaluated on domain $i$ immediately after training, and is evaluated on every domain seen so far, building a $T \times T$ matrix of bits-per-byte (BPB) scores (Figure 5).

\begin{figure}[H]
  \centering
  \includegraphics[width=0.75\linewidth]{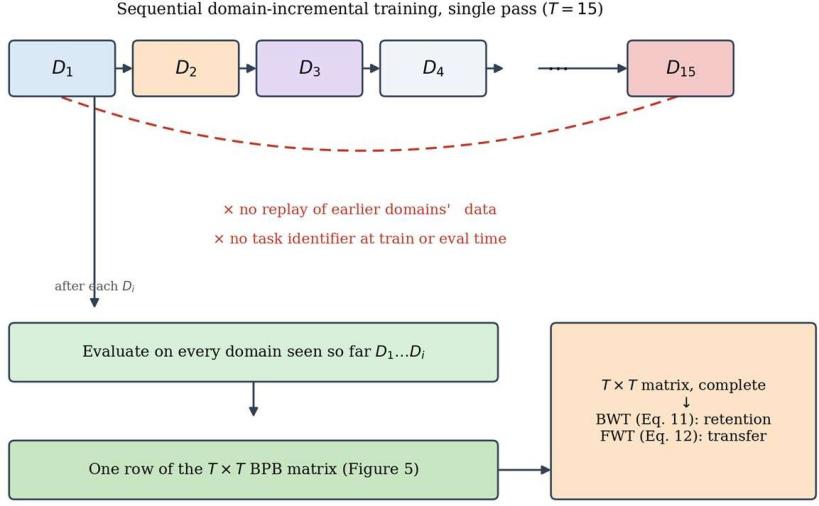}
  \caption{Domain-incremental training protocol: sequential single-pass training with no replay and no task ID, evaluated against every prior domain after each step to build the BPB matrix.}
  \label{fig:protocol}
\end{figure}

The takeaway: no replay and no task ID are both real constraints, not conveniences, so anything the model retains has to come from the architecture itself, not from being told which domain it is in or getting to see old data again.

We report Backward Transfer (BWT) and Forward Transfer (FWT), adapted from Lopez-Paz and Ranzato~\citep{lopezpaz2017gradient} to BPB (lower is better, so the sign convention is inverted from the original accuracy-based formulation):

\begin{align}
\label{eq:bwt}
\text{BWT} &\;=\; \frac{1}{T-1}\sum_{i=1}^{T-1}\bigl( \text{BPB}(i,\, T) \;-\; \text{BPB}(i,\, i) \bigr) \\[0.2em]
\label{eq:fwt}
\text{FWT} &\;=\; \frac{1}{T-1}\sum_{i=2}^{T}\bigl( \text{BPB}_{0\text{shot}}(i) \;-\; \bigramop(i) \bigr)
\end{align}

\subsubsection*{Sign convention, stated explicitly to avoid the inversion error we made ourselves mid-project:}

Positive BWT indicates net forgetting (BPB got worse); negative BWT indicates genuine backward transfer (BPB improved after learning unrelated domains). Negative FWT indicates the model outperforms the bigram floor on a domain it has not yet been trained on.

The 15 domains (Wikipedia, CPython source, Shakespeare, KJV Bible, Reuters news, Moby Dick, the Federalist Papers, Darwin, Whitman, Lodash source, Marcus Aurelius, Flatland, Alice in Wonderland, Python technical documentation, and a Hindi corpus) form a self-assembled domain-shift corpus, not a recognized benchmark. We use the established BWT/FWT formulation on data we compiled ourselves; this is stated plainly as a limitation (Section 6), not implied to be a standard evaluation.

\subsection{Accuracy audit}
\label{subsec:audit}

\subsubsection*{Question: is there anything real for a forgetting metric to measure here?}

Before reporting forgetting, we verify the model has learned something meaningful to forget: a model that fails to beat a trivial bigram frequency table has no real signal for a forgetting metric to measure. Across all 15 domains and all three seeds (42, 43, 44), CMP's post-training BPB beats the bigram floor with no exceptions, with margins ranging from $+0.41$ to $+1.22$ BPB. We apply the identical audit to every baseline reported below. With that established, the next question is whether the central result replicates or was a single favourable run.

\subsection{Main result: three-seed replication}
\label{subsec:replication}

\subsubsection*{Question: does the weight-protect effect hold up across seeds, or was the first number we saw a fluke?}

\begin{table}[H]
  \centering
  \caption{Three-seed replication, weight-protect configuration, 15-domain sequence.}
  \label{tab:replication}
  \small
  \begin{tabular}{@{}lcc@{}}
    \toprule
    \textbf{Seed} & \textbf{BWT} & \textbf{FWT} \\
    \midrule
    42            & $+0.2353$    & $+0.5956$ \\
    43            & $+0.2352$    & $+0.2978$ \\
    44            & $+0.2512$    & $+0.3175$ \\
    \midrule
    Mean $\pm$ std & $+0.2406 \pm 0.0092$ & $+0.4036 \pm 0.1665$ \\
    \bottomrule
  \end{tabular}
\end{table}

BWT is tight across seeds (std $0.0092$, under 4\% of the mean); FWT is not (std $0.1665$, over 40\% of the mean), driven by seed 42's outlier $+0.5956$ against 43 and 44's closer $+0.30$ range. We report this rather than smoothing it: the backward-transfer result this paper's claim rests on is stable across seeds, but the forward-transfer number should be read as noisy and not over-interpreted. A stable low BWT is not yet evidence of anything relative to prior work; the next step is the actual head-to-head.

\subsection{Baseline comparison: naive fine-tuning and EWC}
\label{subsec:baseline}

\subsubsection*{Question: does CMP actually beat the standard fix for forgetting, at matched capacity and budget, or is this a weak-baseline artifact?}

We compare against a parameter-matched ($\sim$6.5M) byte-level Transformer, under two conditions on the identical protocol: naive sequential fine-tuning (no mitigation) and online EWC~\citep{schwarz2018progress}, a running-Fisher variant of Kirkpatrick et al.~\citep{kirkpatrick2017overcoming}, chosen for tractability across many sequential domains rather than per-task Fisher accumulation, stated explicitly, this is not vanilla EWC.

\subsubsection*{A methodological note was included deliberately:}

An initial Transformer baseline run, at the original $150{,}000$-byte data budget, failed the accuracy audit outright: post-training BPB was worse than the bigram floor on every domain, the result of severe overfitting under $728\times$ data repetition at that budget. The resulting BWT ($+5.93$ naive) reflected training instability, not forgetting, and is not reported as a result. Raising the data budget to 3.8M bytes resolved this; all Transformer results below pass the accuracy audit cleanly.

\begin{table}[H]
  \centering
  \caption{Forgetting comparison, matched architecture size ($\sim$6.5M params) and matched data budget, identical protocol.}
  \label{tab:baseline}
  \small
  \begin{tabular}{@{}lcl@{}}
    \toprule
    \textbf{Method} & \textbf{BWT} & \textbf{Notes} \\
    \midrule
    Naive fine-tune (Transformer)   & $+2.7897$ & No mitigation \\
    Online EWC (Transformer)        & $+2.2457$ & $\sim$20\% reduction vs.\ naive \\
    CMP (weight-protect)            & $+0.1482$ & Same 3.8M-byte budget \\
    \bottomrule
  \end{tabular}
\end{table}

\begin{figure}[H]
  \centering
  \includegraphics[width=0.6\linewidth]{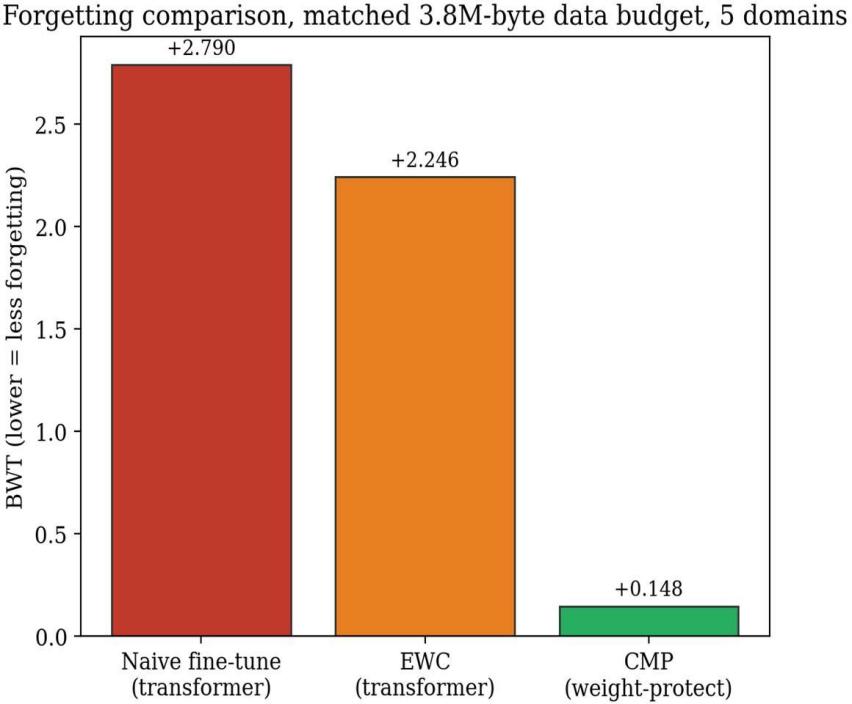}
  \caption{Forgetting comparison, matched 3.8M-byte data budget, 5 domains.}
  \label{fig:baseline}
\end{figure}

\subsubsection*{Reconciling this table with Table 2, explicitly:}

CMP's figure here ($+0.1482$) is not the Table 2 replication mean ($+0.2406$); it is CMP re-run at the 3.8M-byte budget specifically to match what the Transformer baseline required to pass its own accuracy audit (Section 4.2), rather than reusing the Table 2 number, which was measured at the original $150{,}000$-byte budget. This is the same CMP configuration and protocol in both tables; only the data budget differs, and we report the number from the budget that makes the three-way comparison genuinely matched rather than defaulting to whichever figure was more favorable.

CMP forgets approximately 15--19$\times$ less than EWC and naive fine-tuning respectively, on an identical, matched-capacity, matched-budget comparison. This is the paper's central quantitative claim. A single domain ordering, however, is not enough to trust that claim; the next question is whether it survives a different one.

\subsection{Domain-order control}
\label{subsec:order}

\subsubsection*{Question: is the forgetting-resistance result specific to the one domain order we happened to test first?}

This control uses a 5-domain subset (wiki, cpython, shakespeare, kjv, reuters) of the 15-domain main protocol, not the full sequence: reordering all 15 domains across 3 conditions was not computationally tractable within this study, so we test order-sensitivity on a smaller, fully-crossed subset instead and report it as a bound on the effect, not a claim about the full 15-domain ordering.

\begin{table}[H]
  \centering
  \caption{Domain-order sensitivity, matched $150{,}000$-byte budget.}
  \label{tab:order}
  \small
  \begin{tabular}{@{}llc@{}}
    \toprule
    \textbf{Order} & \textbf{Sequence} & \textbf{BWT} \\
    \midrule
    Canonical & wiki $\to$ cpython $\to$ shakespeare $\to$ kjv $\to$ reuters & $+0.2353$ \\
    Reverse   & reuters $\to$ kjv $\to$ shakespeare $\to$ cpython $\to$ wiki & $+0.4442$ \\
    Shuffled  & shakespeare $\to$ wiki $\to$ reuters $\to$ cpython $\to$ kjv & $+0.3343$ \\
    \bottomrule
  \end{tabular}
\end{table}

\subsubsection*{We report this honestly rather than omit it: order matters.}

The canonical ordering is the best case among the three tested, not representative. We therefore report CMP's BWT as a range, $+0.24$ to $+0.44$, rather than the single canonical figure. Even the worst tested ordering still outperforms EWC ($+2.2457$) by roughly $5\times$. We note explicitly that three orderings are a sample, not an exhaustive sweep: five domains admit $5! = 120$ possible orderings, and the true worst case could fall outside the range reported here; we have not searched for it.

\begin{figure}[H]
  \centering
  \includegraphics[width=0.7\linewidth]{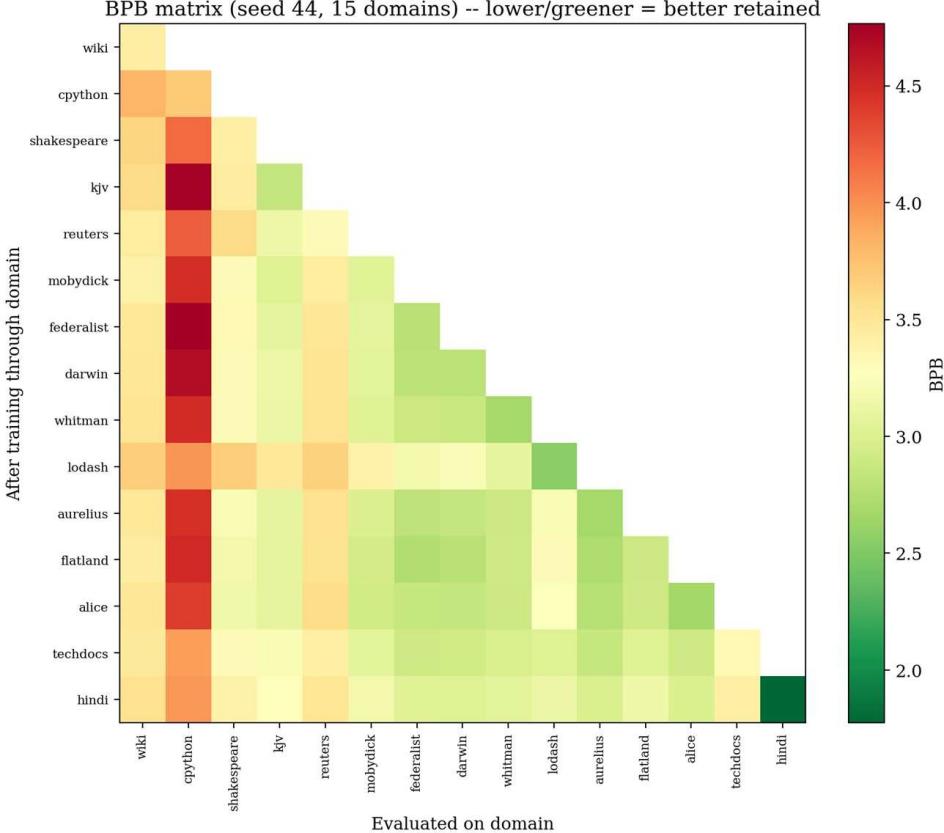}
  \caption{BPB matrix, seed 44, 15 domains; lower/greener is better retained. Note the persistently high (red) cpython column: this domain is retained worse than others across the sequence.}
  \label{fig:bpb-matrix}
\end{figure}

Every result to this point is text. The last open question is whether any of this is specific to text, or whether it is a property of the mechanism that shows up outside it too.

\subsection{Vision anchor: Split-MNIST}
\label{subsec:vision}

\subsubsection*{Question: does the forgetting-resistance mechanism generalize outside text at all?}

To test generalization beyond text on a recognized benchmark, we serialize MNIST digits via a Hilbert curve into the literature-standard 5-task split ($0/1 \to 2/3 \to 4/5 \to 6/7 \to 8/9$).

\begin{table}[H]
  \centering
  \caption{Split-MNIST (5-task), Hilbert-serialized, dense vs.\ sparse (weight-protect), mean $\pm$ std across 3 seeds (42, 43, 44).}
  \label{tab:vision}
  \small
  \begin{tabular}{@{}lcc@{}}
    \toprule
    \textbf{Condition} & \textbf{BWT} & \textbf{FWT} \\
    \midrule
    Dense (base)             & $-0.0211 \pm 0.0004$ & $-0.0669 \pm 0.0013$ \\
    Sparse (weight-protect)  & $-0.0207 \pm 0.0003$ & $-0.0666 \pm 0.0013$ \\
    \bottomrule
  \end{tabular}
\end{table}

\subsubsection*{We report this as a null result, not a success.}

Sparse and dense conditions are nearly identical, and now that we have replicated across three seeds (42, 43, 44) the two conditions' BWT confidence intervals overlap: the difference is within noise, not a real effect. We deliberately do not claim these are ``statistically indistinguishable'' in the formal sense: we have not run a significance test, only a 3-seed mean and standard deviation. We do not use this result to support the forgetting-resistance claim; we include it because disclosing a negative result on a recognized benchmark is more informative than omitting it. This table reports BPB, not classification accuracy, and is therefore not directly comparable to published Split-MNIST accuracy figures for EWC ($\sim$20\%), SI ($\sim$33\%), or replay-based methods ($\sim$90\%+).

\section{Discussion and Limitations}
\label{sec:discussion}

\subsubsection*{Accuracy gap.}
CMP's raw single-domain BPB (3.1--3.27) trails a properly-trained Transformer of matched size by a wide margin. This paper's claim is about the forgetting axis, not raw predictive competitiveness. No non-attention architecture (Mamba, RWKV, RetNet, xLSTM) has cleanly beaten Transformers on short-context accuracy at matched parameters either; a real accuracy gap alongside a real, different contribution is not without precedent.

\subsubsection*{Order sensitivity.}
BWT ranges $+0.24$ to $+0.44$ across three domain orderings, with canonical being the best case, not the average. We report the range throughout. Three orderings are a sample of $5! = 120$ possible sequences, not an exhaustive sweep; the true worst case may fall outside this range.

\subsubsection*{Online EWC, not vanilla EWC.}
Our baseline uses a single running Fisher estimate and anchor~\citep{schwarz2018progress} rather than per-task Fisher accumulation~\citep{kirkpatrick2017overcoming}, chosen for tractability, disclosed explicitly.

\subsubsection*{Custom corpus.}
The 15-domain sequence is self-assembled, not a standardized benchmark. The BWT/FWT formulation is established and published; the underlying data is not.

\subsubsection*{Vision anchor is a null result.}
No measurable advantage of weight-protect over the base configuration on Split-MNIST, replicated across 3 seeds (42, 43, 44); the dense and sparse BWT means sit within one standard deviation of each other. We do not claim vision generality from this experiment.

\subsubsection*{Binding math is not novel.}

The binding operation and memory mechanism are established VSA/associative-memory techniques, not new mathematics. The contribution is the specific combination and the gradient-free plasticity throttle (Eq. 9--10), and the empirical demonstration of a forgetting-resistance property, not a novel mathematical result.

\subsubsection*{Depth is a promising, unresolved direction.}
A separate architecture line reaches substantially lower single-domain BPB (Appendix C). An attempted merge underperformed the flagship configuration (Appendix D); reported as a negative result with a diagnosed, plausible cause, not a closed question.

\section{Conclusion}
\label{sec:conclusion}

We tested a narrow, falsifiable claim: that local, sparse, gradient-free learning resists catastrophic forgetting for structural reasons, not merely when patched with a mechanism like EWC. Across three seeds, a matched-size Transformer baseline, and a domain-order control, CMP shows 15--19$\times$ less backward transfer than online EWC on a controlled domain-incremental protocol. We reported this alongside what does not yet work: a substantial accuracy gap, a null result on Split-MNIST, and a diagnosed failure attempting to combine this architecture with a separate mechanism that improves raw accuracy but has never been tested for forgetting resistance. Closing that gap and a proper scaling-law derivation are the two most direct next steps, both left explicitly open rather than claimed here.

\section*{Appendix}
\addcontentsline{toc}{section}{Appendix}
\renewcommand{\thesubsection}{\Alph{subsection}}
\setcounter{subsection}{0}

\subsection*{A~~Hyperparameters}
\label{app:hparams}

\begin{table}[H]
  \centering
  \caption{Full hyperparameters for the flagship configuration.}
  \label{tab:hparams}
  \small
  \begin{tabular}{@{}ll@{}}
    \toprule
    \textbf{Parameter} & \textbf{Value} \\
    \midrule
    Representation dimension $r$              & 1024 (main result) \\
    Batch size / Sequence length              & 64 / 256 \\
    Steps per domain                          & 6000 \\
    Data budget per domain                    & $150{,}000$ B (Section 4.5) / $3{,}800{,}000$ B (Section 4.4) \\
    Learning rate $\eta$                      & 0.12 \\
    Weight-protect strength $\lambda$         & 5.0 \\
    Sparsity $k$ / buffer / register          & Derived from $r$ at fixed ratio ($64/64/256$ at $r=1024$) \\
    \bottomrule
  \end{tabular}
\end{table}

\subsection*{B~~Per-domain accuracy audit}
\label{app:audit}

Full per-domain, per-seed BPB vs.\ bigram floor for the 15-domain flagship runs (seeds 42, 43, 44) is available in the released result files; every domain, every seed, beats the bigram floor with margins $+0.41$ to $+1.22$ BPB.

\subsection*{C~~Historical depth-line result}
\label{app:depth}

A separate architecture line (stacked, learned multiplicative binding blocks, no memory system, no continual-learning apparatus, pure local one-hop credit) reaches 2.49--2.51 BPB on single-domain byte prediction at 4 blocks, 3-seed replicated, with a shuffled-deepest-block control confirming the gain is causal rather than incidental capacity. This result predates the present architecture and was never integrated with it prior to the attempt described below.

\subsection*{D~~Depth+hierarchy merge attempt (negative result)}
\label{app:merge}

We attempted to combine the depth-line mechanism (Appendix C) with the present flagship architecture, replacing the single byte-level bind with a 4-block stack feeding into hierarchy and memory unchanged. The result underperformed the flagship baseline (3.48 vs.\ 3.27 BPB at matched steps). Our diagnosis: the depth-line's blocks use plain normalization (no kWTA sparsification), while the memory system and every downstream term assume a sparse code; feeding memory retrieval a dense vector likely degraded slot-matching quality. A secondary contributing factor is signal dilution: seven or more readout terms competing for the same global error signal. We report this as a diagnosed negative result and a concrete direction for future work, not a dead end.

\subsection*{E~~Preliminary scale probe}
\label{app:scale}

At the original $150{,}000$-byte data budget, a $5\times$ parameter increase ($1024 \to 2600$ representation dimension, 6.4M $\to$ 29.8M parameters) produced negligible improvement, traced to $728\times$ data repetition starving the larger model of new information. At a corrected, matched 3.8M-byte budget, the same comparison (BWT $+0.1482$ at 6.4M vs.\ $+0.1397$ at 29.8M parameters) shows a real but modest improvement from scale once the data bottleneck is removed, substantially smaller than the improvement from fixing the data budget alone. This is explicitly a preliminary probe, not a scaling-law derivation; a proper grid sweep is left to future work.

\subsection*{Code and Data Availability.}
The domain-incremental text corpus is self-assembled from public-domain and open-source sources (Section 4.1); the Split-MNIST serialization uses the standard MNIST dataset. The training and evaluation code is not publicly released at this time; it is available to researchers on reasonable request to the authors.

\subsection*{Acknowledgments.}
We thank the Google for Startups Cloud Program for compute credits that supported the experiments in this paper, run on NVIDIA A100, T4, and L4 GPUs. We thank Arjit Saxena, founder of Resolute Labs, for mentorship and guidance.

\bibliography{references}

\end{document}